\documentclass[runningheads]{llncs}
\usepackage{graphicx}

\usepackage[T2A]{fontenc} 
\usepackage[russian, english]{babel}
\usepackage[utf8]{inputenc}

\usepackage{amsmath}

\usepackage{color}

\usepackage{cite}

\begin{document}
\title{Machine Translation Models Stand Strong in the Face of Adversarial Attacks}
%
%\titlerunning{Abbreviated paper title}
% If the paper title is too long for the running head, you can set
% an abbreviated paper title here
%
\author{Pavel Burnyshev*  \inst{1} \and
Elizaveta Kostenok* \inst{1,2,3} \and
Alexey Zaytsev\inst{1}}
\authorrunning{P. Burnyshev et al.}
% First names are abbreviated in the running head.
% If there are more than two authors, 'et al.' is used.
%
\institute{Skoltech \and
MIPT \and
IITP RAS}
\maketitle              % typeset the header of the contribution
\begin{abstract}
Adversarial attacks expose vulnerabilities of deep learning models by introducing minor perturbations to the input, which lead to substantial alterations in the output. Our research focuses on the impact of such adversarial attacks on sequence-to-sequence (seq2seq) models, specifically machine translation models. We introduce algorithms that incorporate basic text perturbation heuristics and more advanced strategies, such as the gradient-based attack, which utilizes a differentiable approximation of the inherently non-differentiable translation metric. Through our investigation, we provide evidence that machine translation models display robustness displayed robustness against best performed known adversarial attacks, as the degree of perturbation in the output is directly proportional to the perturbation in the input. However, among underdogs, our attacks outperform alternatives, providing the best relative performance.
Another strong candidate is an attack based on mixing of individual characters.
\end{abstract}
\section{Introduction}

Modern neural machine translation models demonstrate high-quality generated translations, and they are widely used in real-world applications as part of automatic translation systems. For this reason, the robustness and reliability of such models become crucial factors. 

Adversarial attacks, as detailed ~\cite{Xu2020AdversarialAA, reviews_1}, encompass a broad range of techniques aimed at exposing and probing the vulnerabilities of these models. These attacks introduce slight perturbations to the input data, which can, in turn, lead to significant misinterpretations or errors in the output. The aim is to understand the model's weak points and stability under these "attacks".

The core concept of an adversarial attack is not conditioned on the data nature: an attacker tries to significantly change the model output by modifying the input object.
Nevertheless, constructing adversaries for NLP models is complicated due to the discrete structure of the text data ~\cite{samanta2017crafting, belinkov2017synthetic, zhao2017generating}. 
As we can not straightly use derivatives of the loss function, we compute differentiable approximations of metrics ~\cite{zhukov2017differentiable} and derivatives of the adversarial loss with respect to non-discrete token embeddings.
We can use this idea to generate adversarial examples from the embedding space~\cite{fursov2021gradient}
The work~\cite{fursov2022differentiable} goes further in this way by proposing the use of a generative model to make the adversarial attack work.

However, one can spot a common point in significant part of all these articles: they mostly pay attention to models with an output that consists of a single number. 
Nowadays, many use cases for natural language processing models focus on sequence2sequence problems, where both input and output for a model are sequences.
One particular example of such a problem is a classic machine translation. The input, in this case, is a sequence in one language, and the output is a sequence in another language. Our research can help not only investigate vulnerabilities of these models to adversarial perturbations but also provide new insights on the possibility of detecting anomalies and estimating uncertainty for these models. 

Our main contributions to adversarial attacks on machine translation models are:
\begin{itemize}
    \item We propose new techniques to construct adversaries on the machine translation task. The first algorithm replaces input tokens based on the gradient of the target function with respect to the model's embeddings. Another approach exploits approximations of non-differentiable metrics.
    \item We conduct a fair comparison of different attacks based on a diverse set of metrics for a machine translation problem.
    \item Our experiments demonstrate that modern machine translation models are only slightly vulnerable to adversarial inputs. They do degrade for carefully created adversarial examples via a range of techniques, while the effect is less evident compared to drastic performance drops for computer vision and NLP classification models~\cite{2019_evaluate}.
    \item The biggest vulnerability comes from attacks that work at character levels suggesting that in this case the adversarial examples fall out of the domain of data used for training.
\end{itemize}

\section{Related work}

Various types of adversarial attacks on machine translation models have detected their sensitivity to disrupted inputs~\cite{reviews_2, reviews_3}. The first family of attack strategies finds the most loss-increasing perturbations of the source sentence using a gradient in the embedding space. The HotFlip attack~\cite{ebrahimi2018hotflip} vectorizes simple char-level operations such as replacement, deletion, and insertion and uses directional derivatives to select the change of input sample. Targeted attack~\cite{sadrizadeh2023targeted2} uses gradient projections in the latent space to make perturbations. It preserves the similarity between initial and adversarial translations by inserting a target keyword into adversarial output. AdvGen algorithm ~\cite{AdvGen} works on word level and craft adversarial examples based on the similarity between the loss gradient and distance between initial word and adversarial candidates.

The second group of attacks exploits differentiable estimations of standard NLP metrics to control text perturbations. Authors of~\cite{zhukov2017differentiable} propose such approximation to BLEU, and authors of ~\cite{LSDE} use a deep learning model to estimate Levenstein distance between sentences. Dependence on metrics allows selecting perturbations in discrete space more naturally.

The third type of attack can successfully fool a machine translation model by imitating typos or letter emission. Authors of ~\cite{belinkov2017synthetic} add synthetic noise to attacked sentences which includes replacement of letters and varying their order. In addition to swapped characters, distorted inputs can contain emojis and profanity ~\cite{vaibhav19naacl}. 

Several approaches can produce high-quality adversarial examples but require more complicated training and generation processes. GAN-based framework ~\cite{zhao2017generating} operates on a sentence level, and its training process is adapted for the discrete data structure. Authors of ~\cite{Zou2019ARG} propose a reinforcement learning paradigm to generate meaning-preserving examples.

 There are certain methods to evaluate adversarial attacks on NLP data. Attack Success Rate measures the proportion of successful attacks, which reduces twice the BLEU score of adversarial translation compared to initial translation ~\cite{ebrahimi2018adversarial}. Authors of ~\cite{2019_evaluate} propose an evaluation framework for attacks on seq2seq models that focuses on the semantic equivalence of the pre- and post-perturbation input.
  
 In this study, we provide a comparison of principal attack types: gradient-based, synthetic, and metric approximation. Our modifications to existing methods allow both saving the semantic and grammar correctness of adversaries and altering the attacked translation.

\section{Methods}

\subsection{General description of a Machine Translation Model}

The backbone of the majority of modern research and production Translation models is a Transformer model~\cite{vaswani2017attention}. It consists of Encoder and Decoder parts, each of which includes sequential application of a multi-head attention mechanism that forces latent representation of tokens to interact with each other. The encoder of the model maps the input sentence $X = \{x_1, x_2\ldots x_n\}$ into latent representation $Z = \{z_1, z_2\ldots z_k\}$. Decoder likewise translates it into output embedding representation. The decoder output goes into the classification head, which chooses the next token $y_j$, the process repeats until the model generates a special end token. Choice of the next output token $y_{<j}$ depends on input text $X$, hidden representations $z$ and already generated text~$y_{<j}$: 

$$
p_{\theta}(Y | X) = \prod_{j = 1}^m p_{\theta}(y_j | y_{<j}, X, Z), 
$$ 
where $\theta$ are model parameters and $Y = \{y_1, \ldots, y_{k'}\}$. 
The loss function can be defined as $
J(\theta, X, Y) = \frac{1}{n} \sum_{i=1}^n -\log P(y_i | X, \theta)
$.

% In our experiments, we used pre-trained Marian and mBART Transformer models ~\cite{junczys-dowmunt-etal-2018-marian, liu-etal-2020-multilingual-denoising}.

\subsection{Gradient Machine Translation attack}
\label{chap:gradient_mt}

The proposed gradient attack algorithm has a white-box full access to the model's parameters $\theta$, adversarial loss $\mathcal{L}_{a d v}$ we want to minimize, and an input sequence of tokens $X$ that corresponds to a text.
We suppose that for a set of tokens, we have a dictionary of embeddings $\mathcal{V}$.
The model works on the token level, and the number of tokens in the alphabet is $|\mathcal{V}|$. 

The core idea of the attack is inspired by Hotflip~\cite{ebrahimi2018hotflip} attack: we iteratively replace input tokens according to the adversarial loss, calculated with respect to the model's input embeddings $\mathbf{e}$.The new token's embedding would minimize the first-order Taylor approximation of adversarial loss:
$$
\underset{\mathbf{e}_{i}^{\prime} \in \mathcal{V}}{\arg \min }\left[\mathbf{e}_{i}^{\prime}-\mathbf{e}_{i}\right]^{\top} \nabla_{\mathbf{e}_{i}} \mathcal{L}_{a d v}.
$$
$\nabla_{\mathbf{e}_{i}} \mathcal{L}_{a d v}$ means computing gradient with respect to token at position $i$. The subtracted part of the expression does not depend on substitute embeddings, so the optimization problem reduces to 
$$
\underset{\mathbf{e}_{i}^{\prime} \in \mathcal{V}}{\arg \min }\left[\mathbf{e}_{i}^{\prime}\right]^{\top} \nabla_{\mathbf{e}_{i}} \mathcal{L}_{a d v}.
$$
To select the replacement token, we try all possible indexes $i$ and compare their respective difference in loss.

\begin{figure}[t!]
    \centering
    \includegraphics[width = 0.4\textwidth]{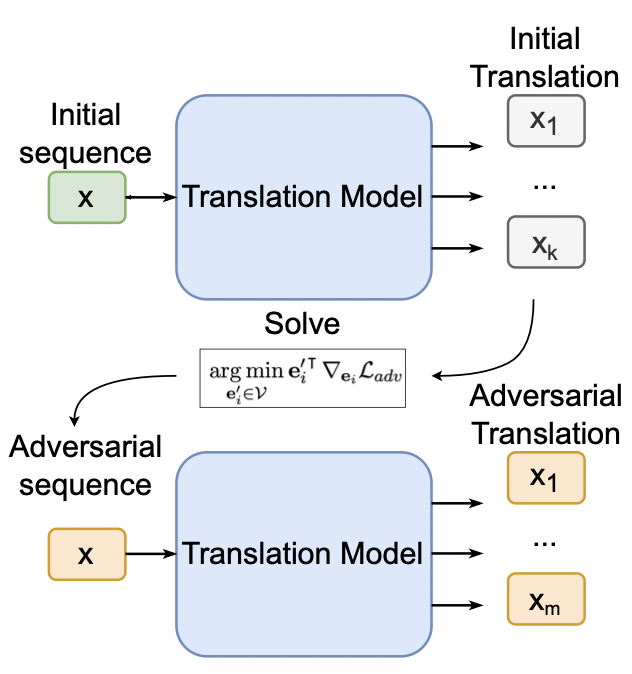}
    \caption{Gradient Attack on Machine Translation models}
    \label{fig:mt_v1_architecture}
\end{figure}

The overall approach is illustrated in Figure~\ref{fig:mt_v1_architecture}.

It's essential to preserve the semantics and grammar of the initial text. Otherwise, the attack discriminators~\cite{zhou-etal-2019-learning} would always detect the attack. So, following~\cite{ebrahimi2017hotflip}, we use several constraints in our experiments. They aim to save the initial meaning of the sentence and prevent an attacker from turning a sentence into a meaningless string of characters that doesn't resemble the initial meaning of a sentence:
\begin{enumerate}
    \item The cosine distance between new and replaced embeddings must not be smaller than the threshold.
    \item Attacker can replace each token position only once.
    \item Attacker can separate all tokens in the vocabulary into two parts. The first part of the subset of tokens always stays at the beginning of the word. Another part stays at the second and next positions. We discourage the algorithm from replacing tokens from one part with tokens from another.
    \item We disallow replacement of tokens denoting punctuation, first and last tokens of the sentence, and stop words. 
\end{enumerate}

\subsection{BLEUER attack}
\label{chap:bleuer_mt}

\begin{figure}[t!]
    \centering
    \includegraphics[width = 0.5\textwidth]{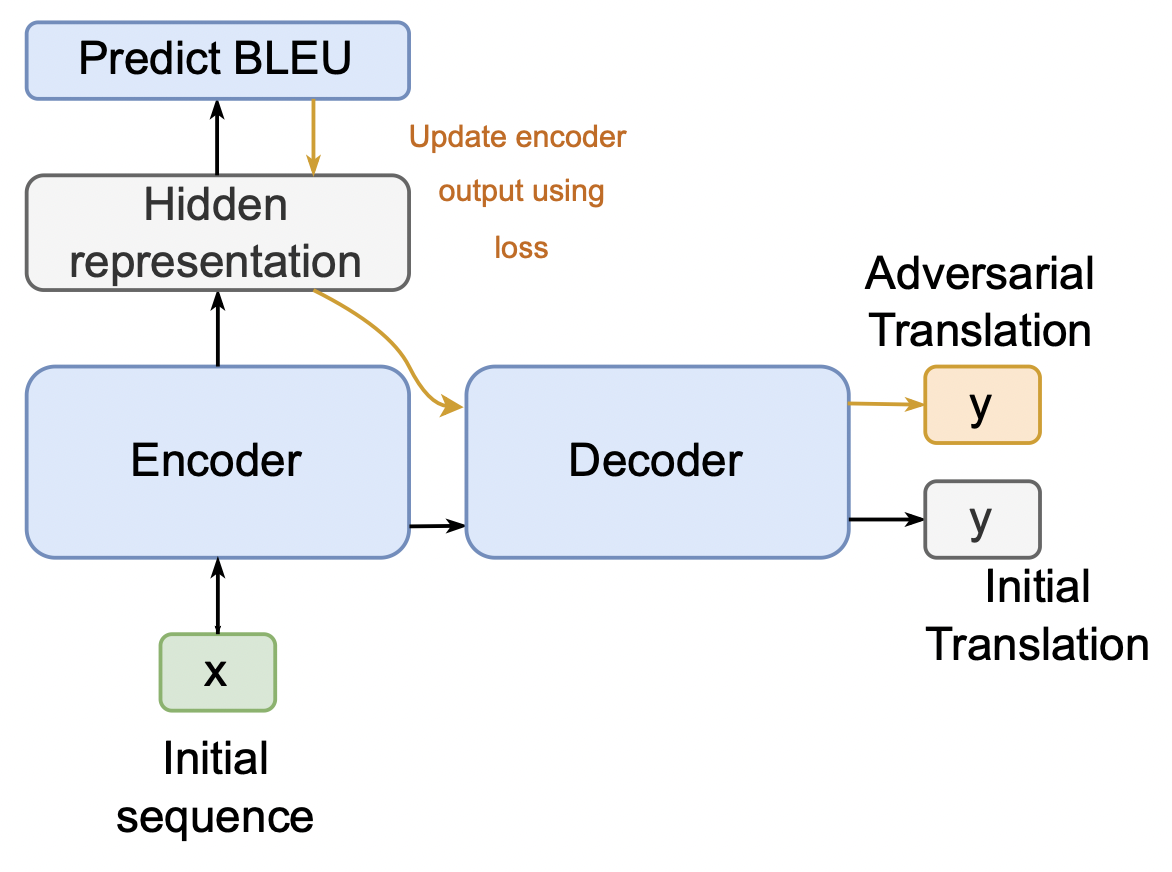}
    \caption{BLEUER Attack, based on predicting initial BLEU score}
    \label{fig:mt_v2_architecture}
\end{figure}

  The gradient attack described before does not guarantee any estimations on the primary translation metrics change: BLEU, METEOR, etc. Instead of optimizing adversarial loss, which does not directly depend on text metric, we can incorporate dependence from the differentiable approximation of the metric. An illustration of the approach for BLEU score is presented in Figure~\ref{fig:mt_v2_architecture}. 

Before applying an attack, an adversary needs to train extra layers to predict BLEU. We translated a subset from the text corpus using the initial translation model and computed initial BLEU scores for pairs of sequences. Those scores are used as targets during the training of additional layers on the top of the encoder part of the model. During the training, we minimize MSE loss between the predicted value of the BLEU score and the original one: 
\[
J = \mathrm{MSE}(f(z), \mathrm{BLEU}(Y_{orig}, Y_{trans})), 
\]
where $z$ is the encoder output, $f$ means applying additional layers, $Y_{orig}$ is expected translation from data corpus, $Y_{trans}$ is the model's translation. 
The attack algorithm executes the following steps: 
\begin{enumerate}
    \item Get encoder outputs $z$;
    \item Calculate prediction of BLEU score $f(z)$ using differentiable layers and loss value $J$;
    \item Calculate gradients with respect to loss and update encoder outputs $z$, so that approximate BLEU score decreases: 
    \[
    z := z + \varepsilon \cdot \nabla_{\mathbf{e}_{i}} \mathrm{MSE}(f(z), 1)).
    \]
    The updated encoder output is served at the entrance to the decoder part of the model, which generates adversarial translation.
\end{enumerate}

\subsection{MBART attack}
\label{chap:mbart_mt}

The proposed gradient attack and BLEUER attack approaches can be combined in the attack, which we call the MBART attack. Depending on gradients, obtained after predicting BLEU score on encoder outputs, we iteratively replace input tokens. After several iterations we get adversarial input $X_{adv}$ and use the model to get adversarial translation $Y_{adv}$.

\subsection{Synthetic attacks}
\label{chap:synthetic_mt}

We propose an extremely naive synthetic attack as an alternative method to attack machine translation tasks. Synthetic attack~\cite{belinkov2017synthetic} initially simulates errors or mistakes that can happen during daily usage of translation systems: keyboard typos, accidental omission/addition, or swapping chars. Additionally, we tried some uncommon sentence perturbations, such as randomly swapping a subset of words or randomly swapping a subset of chars in one word. As the main hyperparameter of such an attack, we used a portion of perturbed words or chars in the sentence.

\section{Experiments}
\label{chap:experiments}

We perform the experiments with Marian and MBART Transformer models. 
For them, the comparison is between the three approaches described above: gradient, BLEUER, and synthetic attacks, since each of them represents a principal type of attack method. We also conduct a comparison of existing and novel attack algorithms. We pay special attention to balancing the trade-off between preserving the original sentence and altering the attacked translation. The attack should not be easily recognized by adversarial detectors, so the text should save logical and semantic literacy and grammar structure. 
We use a wide range of automatic linguistic metrics to evaluate attack approaches from this point of view. The code of our experiments will be available on a public online
repository in the case of acceptance.
% \footnote{\href{https://github.com/fursovia/skindler}.

\subsection{Metrics}

Machine translation attacks aim to decrease the quality of translation metrics. We used $6$ metrics in our experiments.
\textbf{BLEU} is the most famous metric for evaluating the similarity of two sentences, and it is highly correlated with the human concept of text similarity.
\textbf{chrF} metric is calculated as an F-score between character $n$-grams~\cite{popovic-2015-chrf}.
\textbf{METEOR} is another $n$-gram metric. It is calculated as an F-score for unigrams.
\textbf{WER} metric considers a number of basic text operations: adding, deleting, and swapping characters for transforming one text into another.
\textbf{Paraphrase similarity} metric is built upon pre-trained Sentence-Trans-former~\cite{reimers-2019-sentence-bert}, model, which matches texts into $768$-dimensional vectors. Cosine distance between such vectors correlates greatly with a human opinion of text similarity.
\textbf{BertScore}~\cite{zhang2019bertscore} leverages vectors, obtained from pre-trained models. Bert Score has been found to correspond with human judgment at the sentence level meaning. It calculates each token's precision, recall, and F1 measures in assessed sentences. 

\subsection{Baselines}
In addition to the proposed methods, we evaluated naive approaches and state-of-the-art approaches.

In particular, we consider our variant of \textbf{Gradient} attack itself and a modification of it \textbf{Gradient attack and ML constraint}.
For the later attack, we utilize constraints on how much we can change the initial sentence.
These constraints are described above in the methods section and aim at keeping the meaning and structure of the attacked sentence similar to the initial one.
The attack uses Marian~\cite{junczys-dowmunt-etal-2018-marian} Transformers, pre-trained on English-Russian  corpora. 

We consider two types of attacks that consider an approximation of the target metric during an attack: \textbf{BLEUER} and \textbf{MBART} attacks. 
For these attacks, we train an additional head that takes encoder outputs as an input. 
These heads are predicting BLEU or BertScore correspondingly.
As these heads are differentiable, we incorporate these scores into the loss function to maximize the difference between the initial translation $Y$ and the attacked sentence translation $Y_{\mathrm{attacked}}$ and minimize the difference between the initial sentence $X$ and its adversarial perturbation $X_{\mathrm{attacked}}$.
For training of BLEUER and MBART, we use the validation data of \textbf{wmt-14} dataset.

To make sure that all main types of attacks are considered, we evaluate methods from the literature. 
\textbf{Prefix attack} inserts tokens at the beginning as we try to select tokens that serve as a prompt.
\textbf{SWLS} is the attack from the article~\cite{zhang2021crafting}.
This attack tries to leverage a bidirectional translation model and looks for perturbations that maximize the difference between adversarial sequence $X_{\mathrm{attacked}}$ and its back-translation.  

The last two methods consider attacks at separate character levels.
\textbf{Char swap} tries to randomly swap characters to make the attack stronger.
\textbf{Char + grad swap} is a version of our gradient attack at the character level. 

\subsection{Attack examples}

At first, we visually examined the results of the conducted attacks by comparing examples of adversarially perturbed sentences. 
While sometimes the results are imperfect, in general, we see the desired effect.
Examples of such sentences are provided in Table~\ref{table:attack_samples_mt_grad}.
% Appendix~\ref{app_b}.

\subsection{Experiment Setup}
\label{chap:exp_setup}

For gradient attack~\ref{chap:gradient_mt} and "BLEUER" attack~\ref{chap:bleuer_mt} we used Marian~\cite{junczys-dowmunt-etal-2018-marian} Transformers, pre-trained on English-Russian text corpora. For MBART~\ref{chap:mbart_mt} attack we used MBart-50~\cite{DBLP:journals/corr/abs-2008-00401}). For "BLEUER" we additionally trained layers for approximating actual BLEU metric. For training we used validation data of \textbf{wmt-14} dataset.

\begin{table*}[t!]
\centering
% \begin{tabular}{p{1.5cm}p{3cm}p{10.5cm}}
\begin{tabular}{p{1.5cm}p{3cm}p{7.5cm}}
\hline
Attack type & Sentence type & Sentence \\ \cline{1-3} 
Gradient &  Orig. sentence & Cars get many more miles to the gallon. \\   \cline{2-3} 
&         Attacked sentence &  Cars get many more miles to the \textcolor{red}{ormoneon}. \\   \cline{2-3} 
&         Orig. translation & Автомобили проезжают больше миль на один галлон. \\   \cline{2-3} 
&         Translation & Машины проехали еще много миль до галлона. \\   \cline{2-3} 
&         Attacked translation & Машины проехали гораздо больше миль до гормона. \\ 

\hline

BLEUER &    Orig. sentence & Cars get many more miles to the gallon. \\   \cline{2-3} 
&         Attacked sentence & Cars get many more miles to the \textcolor{red}{gall }on. \\   \cline{2-3} 
&         Orig. translation & Автомобили проезжают больше миль на один галлон. \\   \cline{2-3} 
&         Translation & Автомобили проезжают гораздо больше миль до галлона. \\   \cline{2-3} 
&         Attacked translation & Автомобили получают гораздо больше миль до галлона. \\   \cline{1-3}

\hline

Synthetic &         Orig. sentence & Cars get many more miles to the gallon. \\   \cline{2-3} 
&         Attacked sentence & \textcolor{red}{arCs egt myna emro ielsm to het gllnoa.} \\   \cline{2-3} 
&         Orig. translation & Автомобили проезжают больше миль на один галлон. \\   \cline{2-3} 
&         Translation & Машины проехали еще много миль до галлона. \\   \cline{2-3} 
&         Attacked translation & arCs eggt myna empro ielsm to het glnoa. \\   \cline{1-3}

\hline
\end{tabular}
\caption{Attack samples for the Machine Translation task for different types of attacks}
\label{table:attack_samples_mt_grad}
\end{table*}

\subsection{Main results}

There is an important factor to be considered while evaluating machine translation adversarial attacks: perturbations should preserve the lexicon and grammar structure of the initial sentence. Authors of~\cite{2019_evaluate} proposed a new definition, \textit{meaning-preserving} perturbations, which underline the importance of the correct assessment of an attack. We decided to take care of the balance of perturbing initial sentences and translations. Computing two similarities between the source sentence $X$ and its perturbed sentence $X_{\mathrm{attacked}}$ and the similarity between the initial translation $Y$ and translation of the attacked sentence $Y_{\mathrm{attacked}}$ is key to holding such a balance. Suppose the violation of the initial sentence approximately coincides with the violation of the translation. In that case, we cannot talk about the attack's success: the model honestly works out on a distorted sentence. An ideal attack would slightly change the initial similarity metric but significantly decrease the similarity between translations. 

We vary the ability to introduce modifications into the initial sentence by modifying hyperparameters for all attacks. 
For each attack setting, they form a Pareto frontier, which can help us analyze the attack's impact.
Numbers near the dots indicate hyperparameters of the attack. For the gradient attack and BLEUER attack, we provided a threshold for minimum cosine distance between vectors of original and substitute tokens for each dot. For synthetic attacks, we provided a maximum number of basic transformations for each dot.

\begin{figure}[t]
    \centering
    \includegraphics[width = 0.5\textwidth]{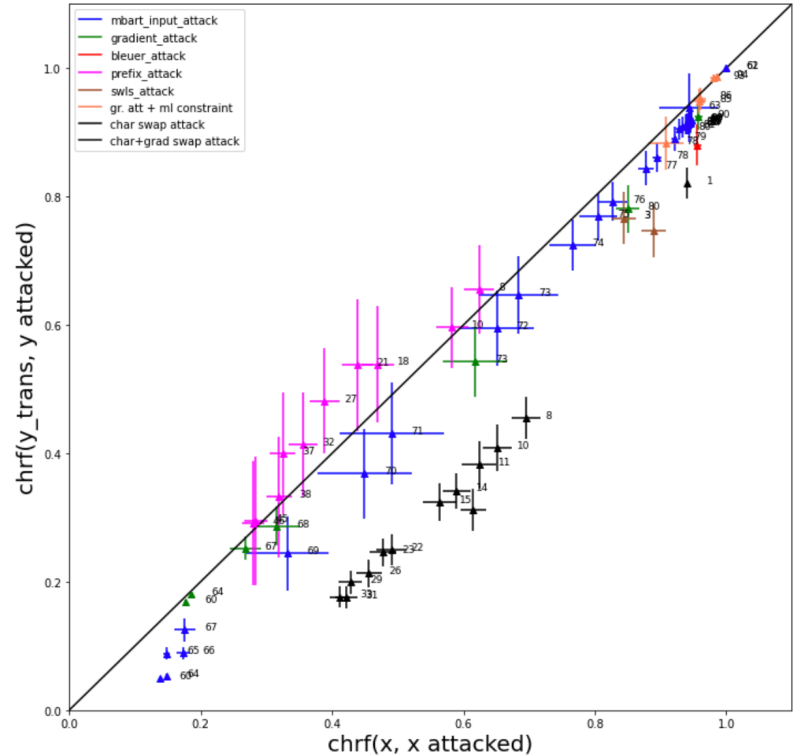}
    \caption{Pareto frontiers for ChRF metric for considered attack methods. We aim at the lower right corner with high change of the translated sentence, but small change of the sentence to translate}
    \label{fig:attack_baselines}
\end{figure}

Pareto frontiers for the full set of considered attacks are presented in Figure~\ref{fig:attack_baselines}. 
In general, the considered attacks could not show high values of attack success rates, supporting the evidence that modern translation models are robust due to the architectural features of models, computational expenses on training, and the colossal size of datasets. The top-performing attack is based on a swap at the level of characters.
Both modifications show a significant improvement over others jointly providing a desired Pareto frontier.

\subsection{Performance with respect to different metrics}

\begin{figure}
    \includegraphics[width=.45\textwidth, height=5 cm]{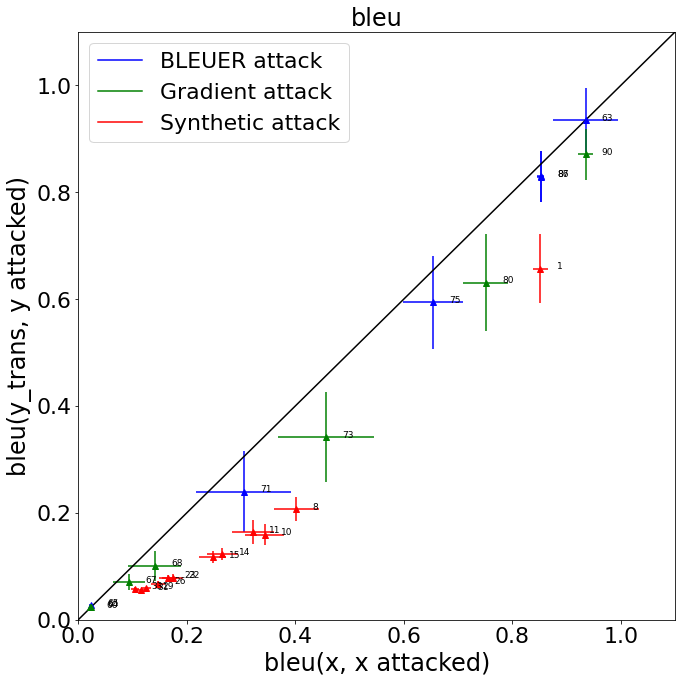}\hfill
    \includegraphics[width=.45\textwidth, height=5 cm]{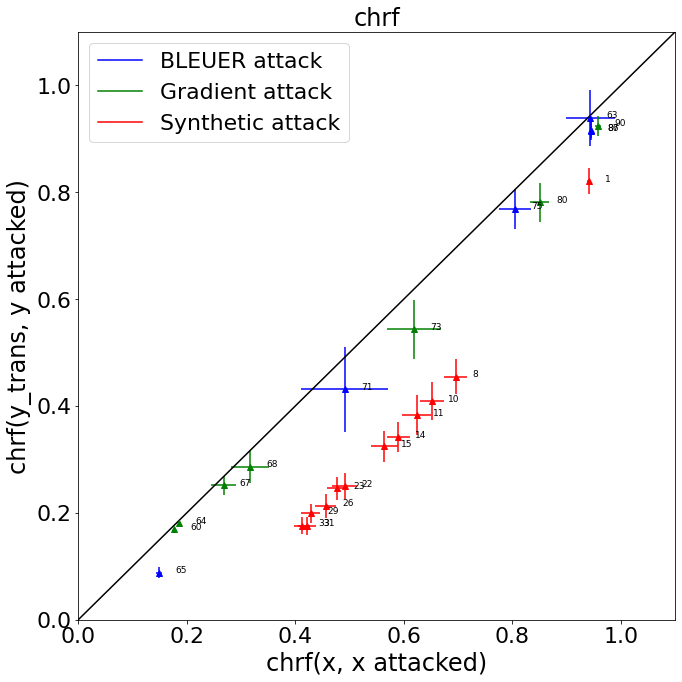}
    \\[\smallskipamount]
    \includegraphics[width=.45\textwidth, height=5 cm]{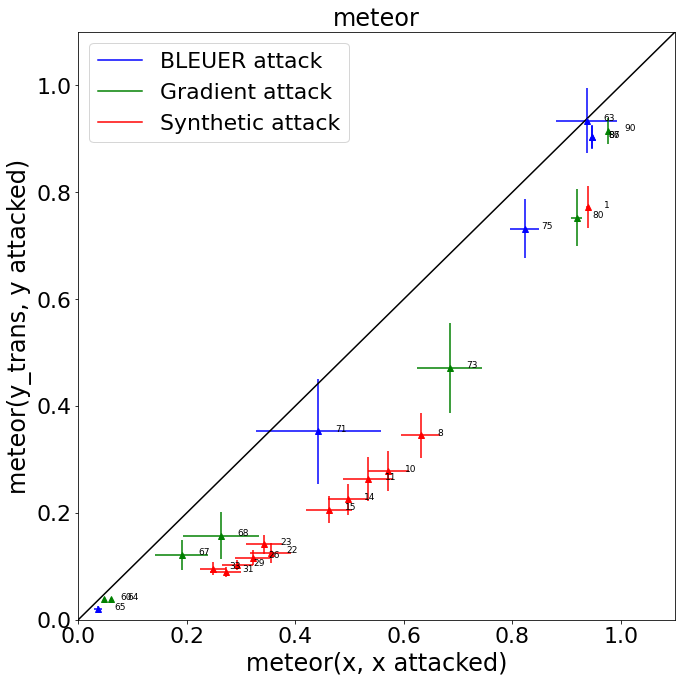}\hfill
    \includegraphics[width=.45\textwidth, height=5 cm]{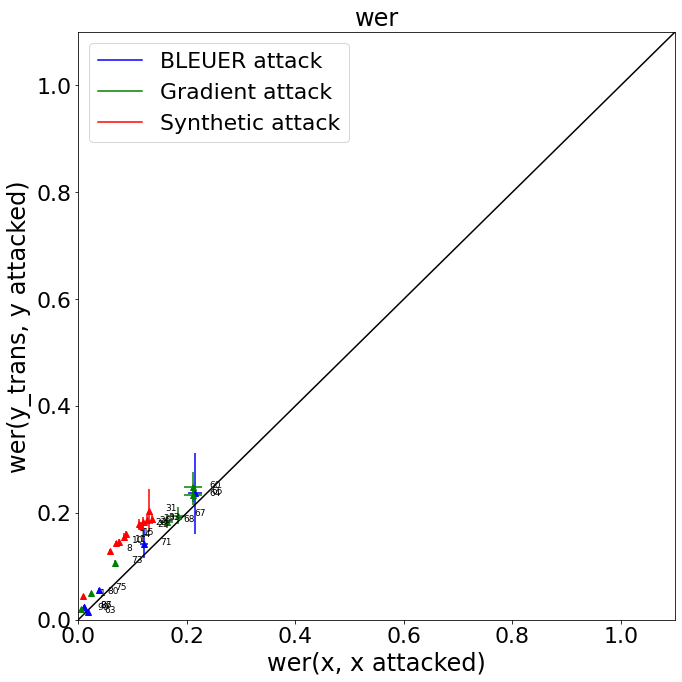}
    \\[\smallskipamount]
    \includegraphics[width=.45\textwidth, height=5 cm]{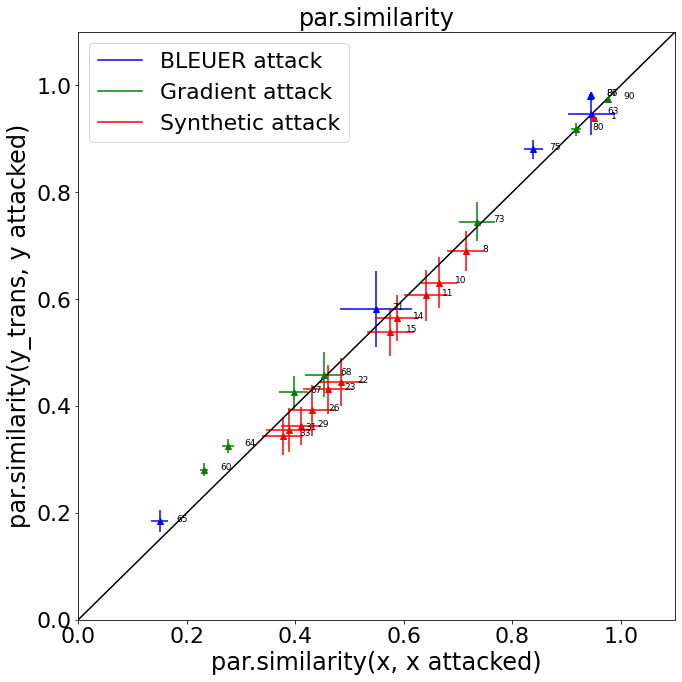}\hfill
    \includegraphics[width=.45\textwidth, height=5 cm]{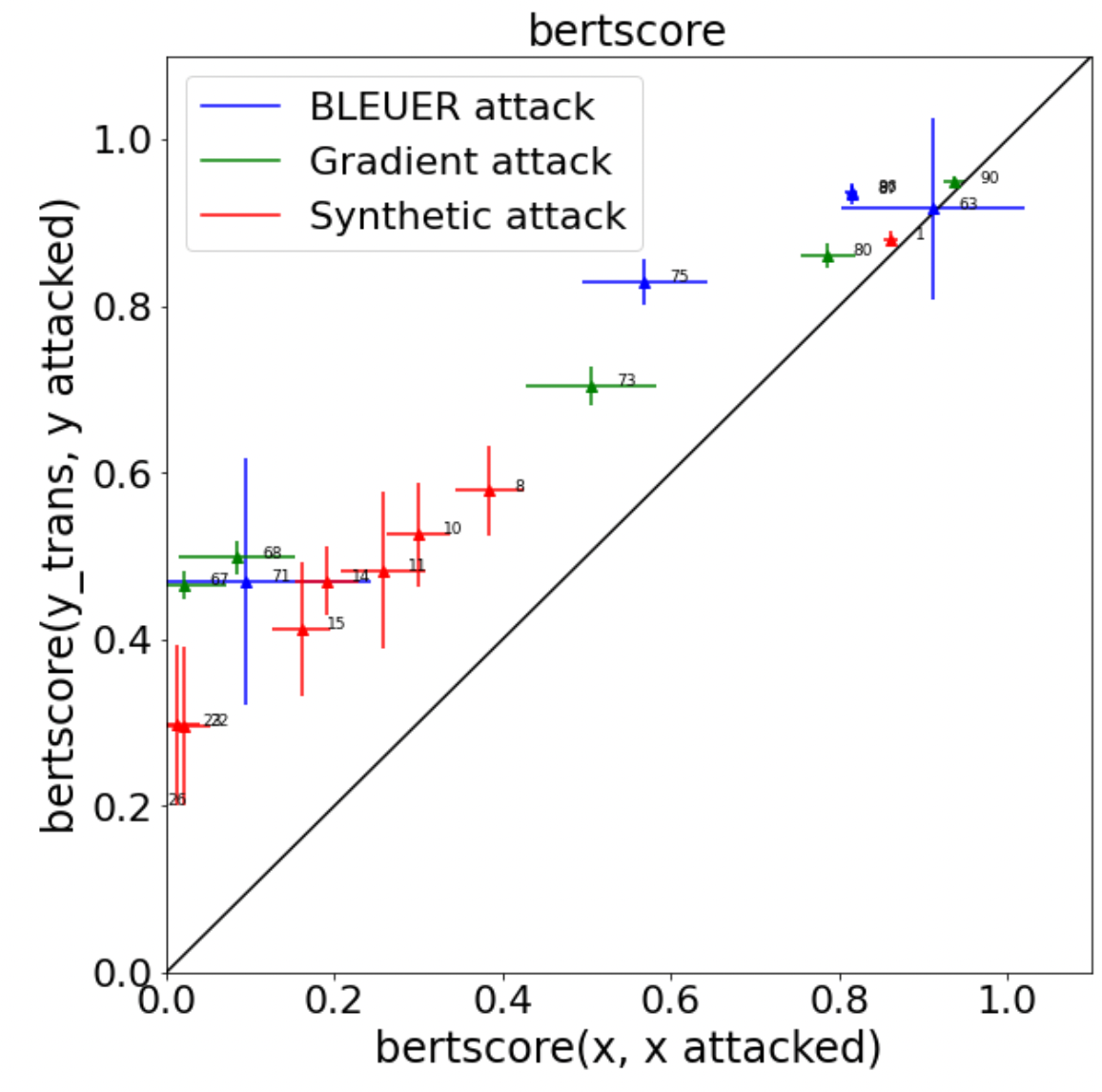}
    \caption{Pareto frontiers for BLEU, ChRF, METEOR, WER, Par.Similarity, BertScore metrics for different attacks. Better attacks should aim lower right corner with a big similarity between input sequences before ($x$) and after an attack ($x_{\textrm{attacked}}$) and low similarity between translated sequences before ($y$) and after ($y_{\textrm{attacked}}$)}
    \label{fig:mt_images}
\end{figure}

We provide Pareto frontiers for $6$ automatic text metrics for $3$ types of attacks: gradient attack, BLEUER, and a synthetic attack. 
Experimental results are presented in  Figure~\ref{fig:mt_images}.
Hitting as low and to the right as possible is the most successful attack, showing min distance between original and adversarial sentences and maximum distance between original and adversarial translations. It is rather evident from the graphics that most dots correspond to the same distortion of the initial and translation sequences. 
We can not ignore that the dots of the most straightforward method, synthetic attack on average, lie lower than the dots of more complicated approaches. 
That fact is especially noticeable for chrF metric due to a char-level of that attack. 
Simple char operations break the structure of tokens, heavily damaging deep models for machine translation, which usually work on the token level.
The numerical summary is given in Table ~\ref{table:deltas}. 
It also supports the evidence that Synthetic attack provides superior metrics compared to an embedding-based approach that leverages the gradients of a model.
% Another option is that the presented metrics also can be an imperfect reflection of the true power of adversarial attacks.

\begin{table}[t!]
\centering
\begin{tabular}{l  l  l  l}
\hline
Metric type & BLEUER & Gradient & Synthetic \\ \cline{1-4} 
BLEU $\uparrow$ &  0.08  & 0.11  &  \textbf{0.20} \\ 
chrF $\uparrow$ & 0.09 & 0.13 & \textbf{0.25} \\
METEOR $\uparrow$ & 0.14 & 0.24 & \textbf{0.27} \\   
WER $\downarrow$ &    -0.02 & -0.04 & \textbf{-0.07} \\  
Paraphrase similarity $\uparrow$ & -0.01 & 0.00 & \textbf{0.06} \\  

BertScore $\uparrow$ & \textbf{0.00} & -0.01 & -0.01 \\   \cline{1-4} 

\end{tabular}
\caption{Numerical comparison of the best attack settings based on the differences between the initial similarity metric and the similarity of translations}
\label{table:deltas}
\end{table}

\section{Acknowledgements}

The research was supported by the Russian Science Foundation grant 20-71-10135.

\section{Conclusion}

Adversarial attacks face limitations in the NLP domain. Especially for the machine translation task, both creating adversarial sequences and evaluating attacks become non-trivial. Most of the existing approaches have a high attack success rate, but they still suffer from lacking semantics and losing lexicon and grammar correctness. In our investigations, we focus on how we can make attacks more meaningful and valuable in analyzing the translation model's vulnerabilities. 
We tried to control translation metrics directly by using differentiable approximations.

The primary outcome of MT experiments is that we still did not find a method that guaranteed that translation would be changed stronger than a source sentence. We compared a range of metrics between initial and corrupted sentences and between initial and attacked translations. 
%The synthetic attack resulted in a $0.2$ drop for chrF and METEOR metrics and outperformed more complicated methods.
We made many additional rules and constraints which forced the attack algorithm not to collapse the initial sentence and save its semantic meaning totally, but they did not significantly change the situation.

%
% ---- Bibliography ----
%
% BibTeX users should specify bibliography style 'splncs04'.
% References will then be sorted and formatted in the correct style.
%
\bibliographystyle{splncs04}
\bibliography{bibliography, custom, bibliography_more}

\begin{thebibliography}{10}
\providecommand{\url}[1]{\texttt{#1}}
\providecommand{\urlprefix}{URL }
\providecommand{\doi}[1]{https://doi.org/#1}

\bibitem{belinkov2017synthetic}
Belinkov, Y., Bisk, Y.: Synthetic and natural noise both break neural machine
  translation (2017)

\bibitem{reviews_2}
Chakraborty, A., Alam, M., Dey, V., Chattopadhyay, A., Mukhopadhyay, D.:
  Adversarial attacks and defences: {A} survey. CoRR  \textbf{abs/1810.00069}
  (2018), \url{http://arxiv.org/abs/1810.00069}

\bibitem{reviews_1}
Chen, J., Tam, D., Raffel, C., Bansal, M., Yang, D.: An empirical survey of
  data augmentation for limited data learning in nlp (06 2021)

\bibitem{AdvGen}
Cheng, Y., Jiang, L., Macherey, W.: Robust neural machine translation with
  doubly adversarial inputs. In: In Proceedings of the 57th Annual Meeting of
  the Association for Computational Linguistics. pp. 4324--4333 (2019)

\bibitem{ebrahimi2018adversarial}
Ebrahimi, J., Lowd, D., Dou, D.: On adversarial examples for character-level
  neural machine translation (2018)

\bibitem{ebrahimi2017hotflip}
Ebrahimi, J., Rao, A., Lowd, D., Dou, D.: Hotflip: White-box adversarial
  examples for text classification (2017)

\bibitem{ebrahimi2018hotflip}
Ebrahimi, J., Rao, A., Lowd, D., Dou, D.: Hotflip: White-box adversarial
  examples for text classification. In: Proceedings of the 56th Annual Meeting
  of the Association for Computational Linguistics (Volume 2: Short Papers).
  pp. 31--36 (2018)

\bibitem{fursov2022differentiable}
Fursov, I., Zaytsev, A., Burnyshev, P., Dmitrieva, E., Klyuchnikov, N.,
  Kravchenko, A., Artemova, E., Komleva, E., Burnaev, E.: A differentiable
  language model adversarial attack on text classifiers. IEEE Access
  \textbf{10},  17966--17976 (2022)

\bibitem{fursov2021gradient}
Fursov, I., Zaytsev, A., Kluchnikov, N., Kravchenko, A., Burnaev, E.:
  Gradient-based adversarial attacks on categorical sequence models via
  traversing an embedded world. In: Analysis of Images, Social Networks and
  Texts: 9th International Conference, AIST 2020, Skolkovo, Moscow, Russia,
  October 15--16, 2020, Revised Selected Papers 9. pp. 356--368. Springer
  (2021)

\bibitem{LSDE}
Gomez, L., Rusinol, M., Karatzas, D.: Lsde: Levenshtein space deep embedding
  for query-by-string word spotting. pp. 499--504 (11 2017).
  \doi{10.1109/ICDAR.2017.88}

\bibitem{junczys-dowmunt-etal-2018-marian}
Junczys-Dowmunt, M., Grundkiewicz, R., Dwojak, T., Hoang, H., Heafield, K.,
  Neckermann, T., Seide, F., Germann, U., Aji, A.F., Bogoychev, N., Martins,
  A.F.T., Birch, A.: {M}arian: Fast neural machine translation in {C}++. In:
  Proceedings of {ACL} 2018, System Demonstrations. pp. 116--121. Association
  for Computational Linguistics, Melbourne, Australia (Jul 2018).
  \doi{10.18653/v1/P18-4020}, \url{https://aclanthology.org/P18-4020}

\bibitem{2019_evaluate}
Michel, P., Li, X., Neubig, G., Pino, J.: On evaluation of adversarial
  perturbations for sequence-to-sequence models. Proceedings of the 2019
  Conference of the North  (2019)

\bibitem{popovic-2015-chrf}
Popovi{\'c}, M.: chr{F}: character n-gram {F}-score for automatic {MT}
  evaluation. In: Proceedings of the Tenth Workshop on Statistical Machine
  Translation. pp. 392--395. Association for Computational Linguistics, Lisbon,
  Portugal (Sep 2015). \doi{10.18653/v1/W15-3049},
  \url{https://aclanthology.org/W15-3049}

\bibitem{reimers-2019-sentence-bert}
Reimers, N., Gurevych, I.: Sentence-bert: Sentence embeddings using siamese
  bert-networks. In: Proceedings of the 2019 Conference on Empirical Methods in
  Natural Language Processing. Association for Computational Linguistics (11
  2019), \url{http://arxiv.org/abs/1908.10084}

\bibitem{sadrizadeh2023targeted2}
Sadrizadeh, S., Aghdam, A.D., Dolamic, L., Frossard, P.: Targeted adversarial
  attacks against neural machine translation. In: ICASSP 2023-2023 IEEE
  International Conference on Acoustics, Speech and Signal Processing (ICASSP).
  pp.~1--5. IEEE (2023)

\bibitem{samanta2017crafting}
Samanta, S., Mehta, S.: Towards crafting text adversarial samples (2017)

\bibitem{DBLP:journals/corr/abs-2008-00401}
Tang, Y., Tran, C., Li, X., Chen, P., Goyal, N., Chaudhary, V., Gu, J., Fan,
  A.: Multilingual translation with extensible multilingual pretraining and
  finetuning. CoRR  \textbf{abs/2008.00401} (2020),
  \url{https://arxiv.org/abs/2008.00401}

\bibitem{vaibhav19naacl}
Vaibhav, Singh, S., Stewart, C., Neubig, G.: Improving robustness of machine
  translation with synthetic noise. In: Meeting of the North American Chapter
  of the Association for Computational Linguistics (NAACL). Minneapolis, USA
  (June 2019)

\bibitem{vaswani2017attention}
Vaswani, A., Shazeer, N., Parmar, N., Uszkoreit, J., Jones, L., Gomez, A.N.,
  Kaiser, L., Polosukhin, I.: Attention is all you need (2017)

\bibitem{reviews_3}
Wang, W., Tang, B., Wang, R., Wang, L., Ye, A.: A survey on adversarial attacks
  and defenses in text. CoRR  \textbf{abs/1902.07285} (2019),
  \url{http://arxiv.org/abs/1902.07285}

\bibitem{Xu2020AdversarialAA}
Xu, H., Ma, Y., Liu, H., Deb, D., Liu, H., Tang, J., Jain, A.K.: Adversarial
  attacks and defenses in images, graphs and text: A review. International
  Journal of Automation and Computing  \textbf{17},  151--178 (2020)

\bibitem{zhang2019bertscore}
Zhang, T., Kishore, V., Wu, F., Weinberger, K.Q., Artzi, Y.: Bertscore:
  Evaluating text generation with bert (2019)

\bibitem{zhang2021crafting}
Zhang, X., Zhang, J., Chen, Z., He, K.: Crafting adversarial examples for
  neural machine translation. In: Proceedings of the 59th Annual Meeting of the
  Association for Computational Linguistics and the 11th International Joint
  Conference on Natural Language Processing (Volume 1: Long Papers). pp.
  1967--1977 (2021)

\bibitem{zhao2017generating}
Zhao, Z., Dua, D., Singh, S.: Generating natural adversarial examples (2017)

\bibitem{zhou-etal-2019-learning}
Zhou, Y., Jiang, J.Y., Chang, K.W., Wang, W.: Learning to discriminate
  perturbations for blocking adversarial attacks in text classification. In:
  Proceedings of the 2019 Conference on Empirical Methods in Natural Language
  Processing and the 9th International Joint Conference on Natural Language
  Processing (EMNLP-IJCNLP). pp. 4904--4913. Association for Computational
  Linguistics, Hong Kong, China (Nov 2019). \doi{10.18653/v1/D19-1496},
  \url{https://aclanthology.org/D19-1496}

\bibitem{zhukov2017differentiable}
Zhukov, V., Golikov, E., Kretov, M.: Differentiable lower bound for expected
  bleu score (2017)

\bibitem{Zou2019ARG}
Zou, W., Huang, S., Xie, J., Dai, X., Chen, J.: A reinforced generation of
  adversarial examples for neural machine translation. In: Annual Meeting of
  the Association for Computational Linguistics (2019)

\end{thebibliography}
%
% \appendix
% \include{app}

\end{document}